\documentclass[lettersize,journal]{article}
\usepackage{amsmath,amsfonts}
\usepackage{algorithmic}
\usepackage{algorithm}
\usepackage{array}
\usepackage[caption=false,font=normalsize,labelfont=sf,textfont=sf]{subfig}
\usepackage{textcomp}
\usepackage{stfloats}
\usepackage{url}
\usepackage{verbatim}
\usepackage{graphicx}
\usepackage{multirow}
\usepackage{cite}
\usepackage[colorinlistoftodos]{todonotes}
\usepackage{enumerate,xcolor}

\newcommand{\PreserveBackslash}[1]{\let\temp=\\#1\let\\=\temp}
\newcolumntype{C}[1]{>{\PreserveBackslash\centering}p{#1}}
\newcolumntype{R}[1]{>{\PreserveBackslash\raggedleft}p{#1}}
\newcolumntype{L}[1]{>{\PreserveBackslash\raggedright}p{#1}}

\usepackage{siunitx}

\begin{document}

\title{Difficulty Modelling in Mobile Puzzle Games:\\An Empirical Study on Different Methods to Combine Player Analytics and Simulated Data}

\author{Jeppe Theiss Kristensen \\ 
  A.P. Moller - Maersk \\
  \textit{jeppe.t.kristensen@gmail.com}
    \and 
  Paolo Burelli\footnote{Corresponding Author} \\
  IT University of Copenhagen \\
  \textit{pabu@itu.dk}
}
\maketitle

\begin{abstract}
Difficulty is one of the key drivers of player engagement and it is often one of the aspects that designers tweak most to optimise the player experience; operationalising it is, therefore, a crucial task for game development studios.
A common practice consists of creating metrics out of data collected by player interactions with the content; however, this allows for estimation only after the content is released and does not consider the characteristics of potential future players.

In this article, we present a number of potential solutions for the estimation of difficulty under such conditions, and we showcase the results of a comparative study intended to understand which method and which types of data perform better in different scenarios.

The results reveal that models trained on a combination of cohort statistics and simulated data produce the most accurate estimations of difficulty in all scenarios. Furthermore, among these models, artificial neural networks show the most consistent results.

\end{abstract}

\section{Introduction}

Among the many characteristics of a game that affect the player experience, difficulty is generally considered to be one of the most impactful. As theorised by Csikszentmihalyi regarding the concept of \emph{flow}~\cite{csikszentmihalyi1990flow}, if the game is too easy, players risk not feeling challenged, becoming bored and potentially quitting the game.
On the other hand, in case it is too hard, players might feel stuck, become frustrated, and again, potentially quit the game.
Therefore, it is not surprising that an important focus for many game studios is to be able to estimate the difficulty of game content.

Difficulty estimation is particularly important for live-operated games, which Is currently the largest market in the game industry~\cite{mordorintelligence,newzoo2022,earthweb2022gamingindustry} since new content is constantly produced and old content updated to fit the demand of the players.
Ensuring players \textit{stay} engaged in the game is crucial for these games to be commercially successful, which puts pressure on game studios to have content that is adequately challenging for new players to pique their interest, and also to release new content for more experienced users that can keep the game experience fresh and exciting.
The challenge for any difficulty prediction framework is therefore two-fold: not only should it work on previously released content, where rich information about other players is available but also on novel content where player data is more sparse or non-existent. Furthermore, as the difficulty is a concept that emerges from the interaction between players and content, the estimation is going to be affected not only by the content type but also by the target player or cohort of players.

As a first step in the process of building a difficulty modelling framework, it is necessary to operationalise difficulty so that it can be quantified and measured.
A common gameplay element in many puzzle games is the existence of discrete tasks or levels.
A direct way to quantify the difficulty of a level is by using the average number of attempts the players require to complete the level.
This metric has been linked to player churn and it gives the level designers a clear measure of how much time players are expected to spend on the challenges.
However, the measurement very much depends on the players that have played the level at the time of measurement.
This \textit{perceived} difficulty can vary depending on the cohorts of players and their skill, so using this average metric for all players can be misleading and does not fully inform the level designers about the \textit{intrinsic} difficulty of the level \cite{kristensen2022uist}.

Another challenge for difficulty estimation is the complexity of the gameplay.
Some aspects of a puzzle, such as winning strategies or ``traps'', may not be immediately apparent through a static analysis of the level.
While it may be possible to infer the level's complexity by using historic data of previous players' performances, this is not possible with new content.
Instead, it typically requires multiple level-designers to manually playtest the content to overcome this \textit{cold start} problem, which can be an expensive and time-consuming process.
This limits the volume of new content and also the quality due to the level of the designers' own skills and biases.
A framework for difficulty prediction can therefore benefit from a component that can explore the dynamics of the game to determine the intrinsic difficulty, but also possibly model how individual players respond to the intrinsic difficulty.

A model capable of estimating perceived difficulty in these different scenarios has multiple potential applications: it could be employed to give immediate feedback to designers when working on new puzzles, it could be used to guide a procedural content generation algorithm intended to create content at a specific level of difficulty, or it could be used to adapt content so that it would provide a specific level of challenge to a specific player or cohort.

With these challenges and applications in mind, in this research work, we investigate various difficulty modelling frameworks that can address some of the issues and shortcomings that previous approaches have revealed.
The two main research questions we aim to answer are:

\begin{enumerate}[ \textbf{RQ1:} ]
    \item[\textbf{RQ1:}] Is it possible to accurately estimate the perceived difficulty of a puzzle for a specific player? What combination of model and data is most effective at this task?
    \item[\textbf{RQ2:}] Is it possible to accurately estimate the perceived difficulty of a puzzle for a specific cohort of players? Can we use the same models and methods developed for personalised predictions? What combination of model and data is most effective at this task?
\end{enumerate}

On the one hand, with the first research question (RQ1), our aim is to investigate personalised models of difficulty that can be used for adaptive gameplay. On the other hand, with the second research question (RQ2), we aim to investigate models that are suitable for either cohort-level adaptation or to support procedural content generation and manual content design.

To answer these two questions, we designed two experiments in which we break the questions down and explore how different difficulty prediction methods work in various scenarios.
Using data from a commercial puzzle game, we investigate the impact of using different types of data, methods and granularity for difficulty prediction, which contributes to the body of knowledge for developing a complete and feasible framework for both personalised and cohort difficulty predictions.
While the research builds on results from a puzzle game, the methods and results have the potential to be extended to other types of games with similar characteristics -- i.e. linear progression and discrete repeatable challenges -- that can help game studios and researchers alike for building a more comprehensive model of difficulty in games.

The work described in this article is part of Dr. Kristensen's doctoral thesis and an earlier version of this manuscript can be found in Dr. Kristensen's dissertation~\cite{kristensenPhD2022}.

\section{Related work}

Difficulty modelling and prediction can be approached in many different ways, depending on the use case. 
In this paper, we root our analysis in the practical application of modelling difficulty in a commercial context in which both the player population and the game experience change constantly.
To properly analyse a framework that can work in such a context, we first discuss the definition of difficulty and give an overview of how it has previously been modelled in various contexts.
In addition to that, we also examine how various playtesting methods have been used for evaluating new content and relate that to the approach presented in this paper.

\subsection{Predicting difficulty}

For any measurement of difficulty to be meaningful, it must reflect the experience of the players.
However, there can be multiple aspects that can change this perception, including uncertainty and skill of the player~\cite{denisova_measuring_2020}.
Therefore, it is highly individual how challenging a player finds a puzzle, so a useful notion is \textit{perceived} difficulty, where difficulty is described as a \textit{relational attribute} between the game and the player~\cite{denisova_challenge_2017, kristensen2022uist}.
The importance of the individual perception of difficulty can be understood through the lens of \textit{flow}; a state where the player loses track of time and worries \cite{csikszentmihalyi1990flow,chen_flow_2007}.
To reach this state, the presented difficulty of a task should be optimal in terms of player skill, which means that it should satisfy the players' intrinsic needs to feel competent and lead to an engaging game experience~\cite{Ryan2006,tyack2020self, alexander_investigation_2013}.
With flow and intrinsic motivation being major contributors to player engagement~\cite{brockmyer2009development}, a core design object of the difficulty prediction framework is, therefore, to account for individual or cohort differences.

Pusey et al.~\cite{pusey_puzzle_2021} describe a number of measurable metrics for puzzle games that quantify difficulty, including the number of actions and time taken to complete a puzzle, as well as the number of incorrect attempts.
Indeed, using the average number of attempts players spend to complete the level, or inversely the pass rate, is a common way to operationalise difficulty in puzzle games (e.g. in Angry Birds~\cite{roohi2021predicting}, Lily's Garden~\cite{kristensen2020estimatinglearning}, Candy Crush~\cite{gudmundsson2018human}).
As noted by Denisova et al.~\cite{denisova_measuring_2020}, the perceived difficulty of the player is strongly related to the number of successes and failures; therefore, in this article, we adopt a similar definition of difficulty as the number of attempts to complete a given level.

The objective of predicting the number of attempts has been approached in a number of different ways and with different applications.
One use case is dynamic difficulty adjustment (DDA)~\cite{hunicke2005case,zohaib2018dynamic} where the game content is adjusted so that the perceived difficulty of a player follows an optimal or predefined goal. 
Gonzalez-Duque et al.~\cite{gonzalezduque2021fastmodelling} consider recent play history and use Bayesian methods for estimating how much time a player requires to complete a sudoku puzzle or platform level.
A more direct approach is presented by Xue et al.~\cite{xue_dynamic_2017}, who employ the pass rate associated with different random seeds for each level to serve personalised content. 
Deep learning methods have also been used to model players and optimise difficulty and engagement~\cite{or2021dl, pfau2020enemy}.

Although DDA has the promise of increasing engagement~\cite{constant_dynamic_2019,li_difficulty-aware_2021}, not all games are designed with this kind of dynamic adjustment in mind, and designers instead find it sufficient with more ad hoc or daily predictions.
For this purpose, parametric-based approaches have been applied for estimating the probability of success~\cite{mourato2014difficulty, kreveld2015automated, kristensen2021statistical}.
An example is the work by Wheat et al.~\cite{wheat2016modeling} where four categories of data (e.g. level features and player behaviour) are considered.
In this work, a random forest classifier had the best accuracy when predicting how difficult each player perceived levels in a custom platform game.
However, in more complex games with large, diverse player bases, these parametric approaches may not be able to fully capture the complex inter-relationship between players and levels, leading to limited applicability.
Matrix factorisation methods, known from recommender systems and their ability to deal with sparse data, have been used specifically for predicting the perceived difficulty of each player in games~\cite{kristensen2022uist,zook_temporal_nodate}.

In this work, we build on the prediction methods from Kristensen et al.~\cite{kristensen2022uist} by extending the analysis in two ways.
The main problem is that matrix factorisation methods suffer from a cold start problem where new players or levels are not possible to include during training.
This leads to the model not learning a latent representation of the new users and levels, which in turn makes predictions on new content impossible.
To address this, we experiment with including different categories of data, similar to Wheat et al.~\cite{wheat2016modeling} in addition to testing out both a neural network and a random forest model for difficulty prediction.

\subsection{Playtesting agents}

Although player data and level data have been used for estimating the difficulty of levels, these categories of data do not include any kind of dynamic information about the levels and are bound to be only an approximation of the gameplay.
It is therefore necessary to extract data from playing through the level, and with the complexity of games today, more intelligent solutions than simple heuristic-based methods are increasingly necessary.
The methods have also become mature enough to assist the whole game design process, ranging from finding bugs, edge cases and general playability~\cite{ariyurek2019automated,latos2022automated,dukkanci2021level,bergdahl2020augmenting}, to modelling players and adjusting difficulty~\cite{stahlke2020artificial, kristensen2020estimatinglearning, horn2018monte, roohi2021predicting,pfau2020dungeons}.

In the specific case of puzzle games, (deep) reinforcement learning (RL) and Monte-Carlo Tree Search (MCTS) based approaches are among the most commonly used to estimate the difficulty of levels~\cite{kristensen2020strategies,mugrai2019automated,shin2020playtesting,kamaldinov2019deep,poromaa2017crushing,gudmundsson2018human}.
In one of the most recent examples, Kristensen et al.~\cite{kristensen2020strategies} use an implementation of the RL method Proximal Policy Optimisation (PPO)~\cite{Schulman2017} to develop an agent capable of playing through puzzle levels in the commercial puzzle game Lily's Garden.
Further work showed that despite the agent performing subpar compared to human players, the agent performance is strongly correlated to actual player performance~\cite{kristensen2020estimatinglearning}.
As noted by Zhao et al. \cite{zhao2020winning}, the goal of creating playtesting agents is not necessarily to outperform humans but rather to capture facets of the gameplay related to skill and style that can be related to player behaviour.
With a similar perspective, Gudmundsson et al.~\cite{gudmundsson2018human} demonstrated that by post-processing the statistics produced by an agent playing a commercial puzzle game, it is possible to better capture these facets and produce a more accurate estimation of difficulty.

Inspired by these approaches to agent-based testing and parametric difficulty modelling, in this study, we propose a combined method based on the RL approach by Kristensen et al. \cite{kristensen2020estimatinglearning,kristensen2020strategies}.
The method and its components are chosen and evaluated through a comparative study intended to showcase how different combinations of methods and data perform at the different difficulty estimation tasks described by the two research questions defined in the previous section.
The aim of the study is to conduct a systematic analysis of the difficulty modelling problem in the context of puzzle games and establish a benchmark for future works in the field.

\section{Case study: Lily's Garden}
\label{sec:case study}

\begin{figure}[t]
    \centering
    \frame{\includegraphics[width=.95\columnwidth]{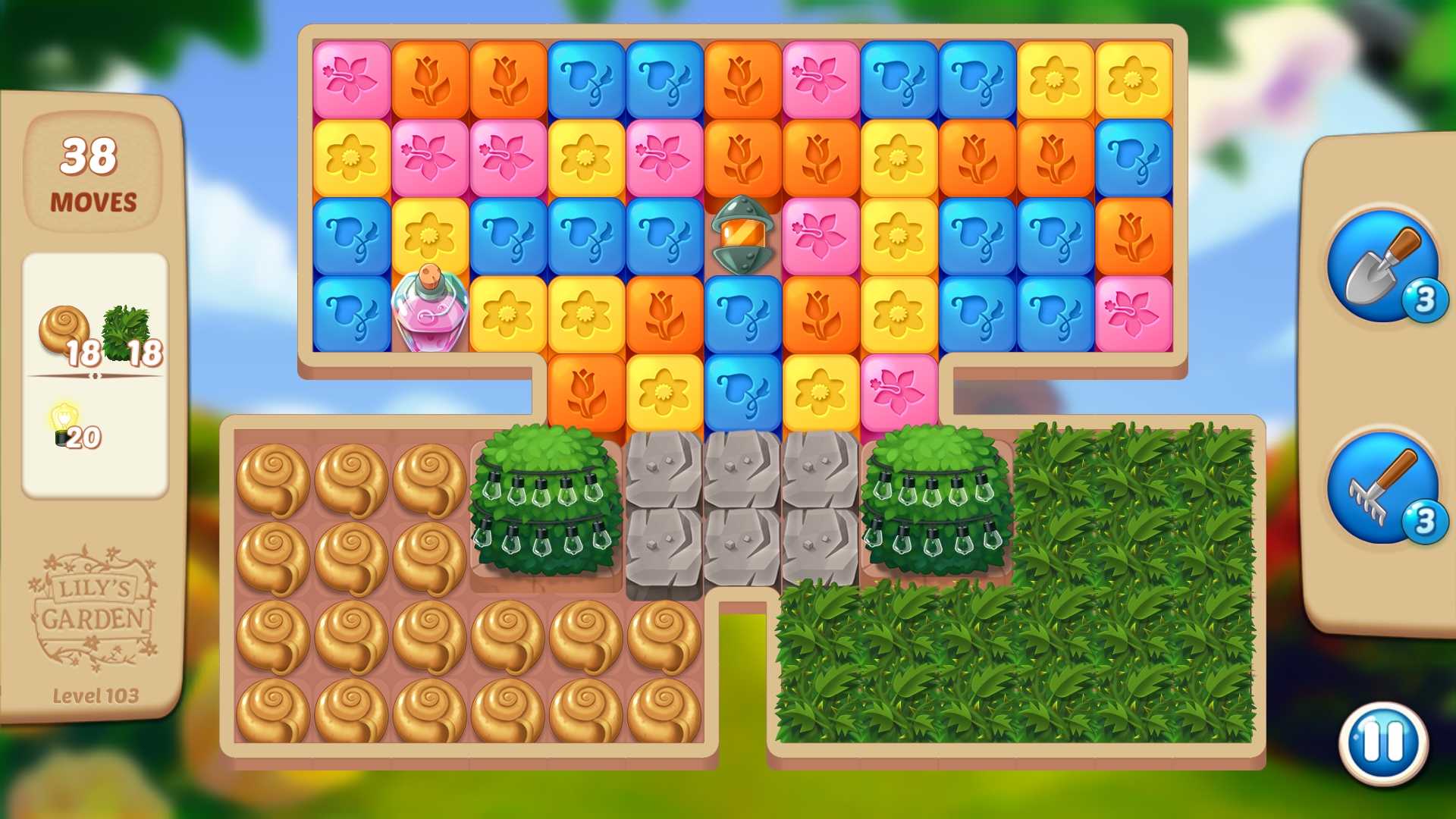}}
    \caption{Example puzzle level in Lily's Garden}
    \label{fig:example level}
\end{figure}

In this article, we employ data from the free-to-play mobile puzzle game Lily's Garden by Tactile Games as a case study to model difficulty in a realistic commercial scenario.
The game was released in early 2019 and has close to a million active daily users worldwide as of today.
It is one of the top 10 grossing puzzle games in the US \cite{appbrain2022topgrossing} and serves as a representative sample of games in this genre both in terms of gameplay and general characteristics.


Lily's Garden is a puzzle game with an overarching narrative in which the player can unlock decorative pieces and story plots by completing puzzle levels.
These puzzle levels are blast-type puzzles and contain a gameboard that can be up to 13 by 9 in size.
An example of a level is shown in Fig. \ref{fig:example level}.
The core gameplay consists of tapping on basic board piece clusters of the same colour to remove the pieces themselves as well as adjacent nonbasic pieces.
By tapping on larger clusters, more powerful pieces can be created, which can clear larger areas on the board.
To complete the level, the player must clear a number of objectives within a given move limit.
As a part of the free-to-play business model the game employs, it is also possible for the player to use booster items that can affect the gameboard directly or add additional moves, which can be acquired from in-game events and purchases.
However, all levels are designed to be completed without the necessity of using any booster.

\begin{figure}[t]
    \centering
    \includegraphics[width=.95\columnwidth]{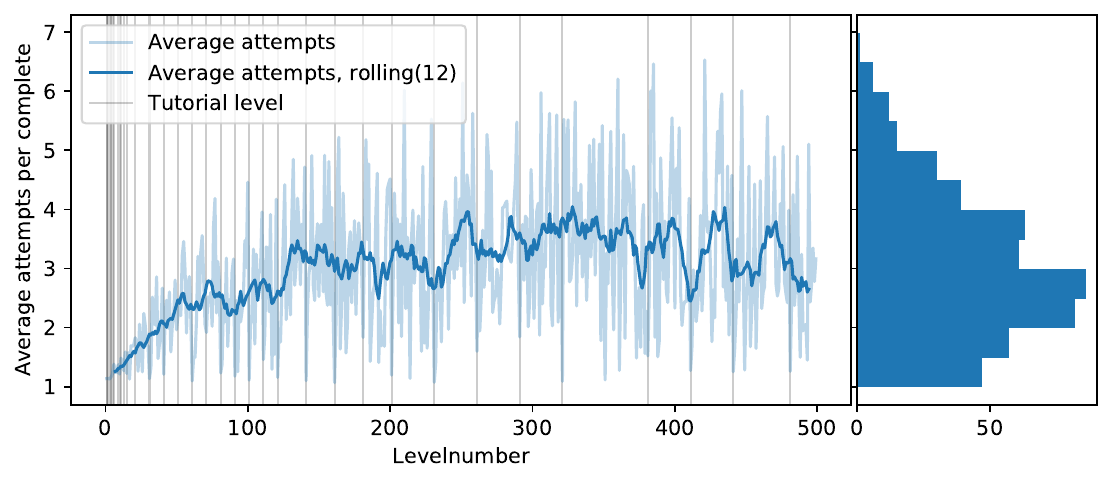}
    \caption{The average attempts per complete over the first 500 levels investigated. A rolling mean with window size 12 is also shown to visualise the trend. The vertical grey bars indicate tutorial levels. Adapted from Kristensen et al. (2022)~\cite{kristensen2022uist}.}
    \label{fig:level range difficulty}
\end{figure}

There are currently more than 6000 available levels, but in this study, we chose to limit the modelling and predicting difficulty to the first 500 levels as they are as a representative sample of the entire level database.
Using the definition of difficulty as the average number of attempts per complete, the difficulty over the level range can be seen in Fig. \ref{fig:level range difficulty}.
The first 100 levels or so contain multiple tutorials and easy levels to properly onboard new players and engage them in the background story.
Subsequently, the difficulty stabilises at around $3.2$ attempts per complete, with easier levels typically being completed in just one attempt and more difficult levels requiring upwards of an average of 7 attempts.

\begin{figure}[t]
    \centering
    \includegraphics[width=.95\columnwidth]{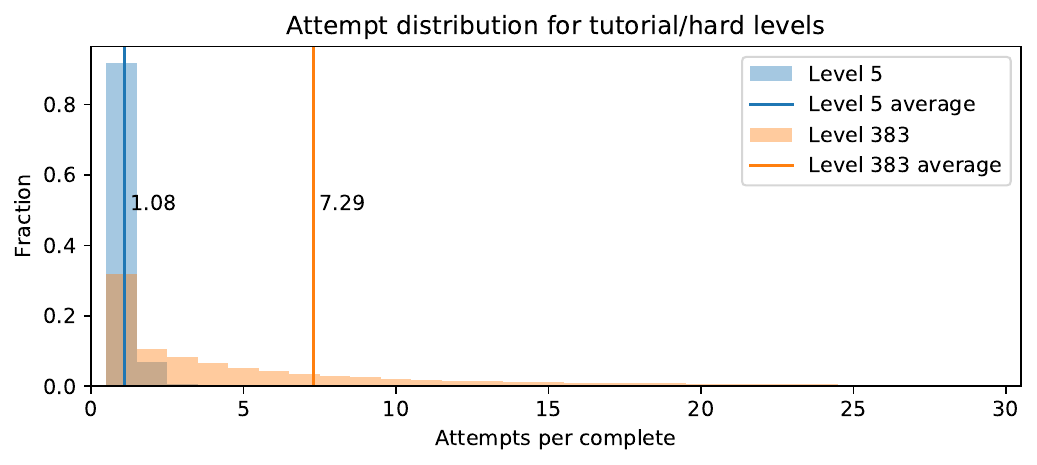}
    \caption{Two examples of attempt distribution on an easy and a hard level (level 5 and 383, respectively). Adapted from Kristensen et al. (2022) \cite{kristensen2022uist}.}
    \label{fig:attempt distribution example}
\end{figure}

While the average varies around 3 attempts, there are large individual differences.
As an example of this, Fig. \ref{fig:attempt distribution example} shows the distribution of attempts on an easy and hard level and their averages.
Players most commonly only spend one attempt on a level but due to both the random elements of the game and player skill, hard levels sometimes require upwards of 30 attempts or more.
This long-tailed distribution appears similar (but not equal) to a geometric distribution where the mean (here the average number of attempts) is proportional to the inverse pass rate and the variance is proportional to the mean and pass rate.

\section{Methods}

In a similar approach as Kristensen et al.~\cite{kristensen2022uist}, we treat the task of predicting the number of attempts a player will spend on a level as a supervised regression task. We employ a two-step approach similar to that used by Gudmundsson et al.~\cite{gudmundsson2018human} where we first extract relevant data from a playtest agent and then train a prediction model. We build on these previous works by testing multiple models and multiple combinations of anylitics and simulated data in different prediction scenarios.
In the following section, we describe the playtest agent used to collect the simulated data and the different types of regression methods that will be used in the experiments.
The specifics and hyperparameters of each set-up are described in the Appendix.

\subsection{Playtest agent}

To explore the dynamics of each puzzle in the game, we employ a reinforcement learning agent where we follow the learnings from previous work on creating a playtesting agent for Lily's Garden \cite{kristensen2020estimatinglearning, kristensen2020strategies}.
In order to capture the game numerically, each level is described by a 3D array of size $13\times9\times n$, where $n=27$ and represents various board piece mechanics.
These mechanics include possible colours, whether the piece is clickable or not and if it is a goal piece.
If a board piece is present in a given position, its hit points are added to the position in each of the channels that match the mechanics of the piece.
In addition to the game board, we also provide information about the number of moves taken, the moves left as well as how many total goal pieces remain after the match.
Possible actions correspond to a position on the game board, meaning that there are $13\times9 = 117$ possible actions, with invalid moves being masked out. A more comprehensive description is given on the website \url{https://aicompetition.tactilegames.com/environment}.

Differently from our previous works aiming to create a single playtesting agent that learns to generalise its strategies to any level, we train one agent for each level to be used as data samples representing user behaviour.
For each level, we collected the data for 8 random seeds, which allows us to test different level configurations and learn multiple optimal strategies for each level.

We let the agent take up to 100 moves per episode, which is around double the average normal move limit of 42. 
In order to encourage the agent to learn to finish the level as quickly as possible, we reward the agent for completing the level and give an additional bonus for completing it before the normal move limit, or a penalty if it spends more than the move limit.
Additionally, we also introduce a small reward for creating power-piece combinations, which can be key to winning some levels.
The final reward function is as follows:
\begin{itemize}
    \item $+0.1$ for each goal collected.
    \item $+0.8$ if completing the level.
    \item $\max(-0.8; +0.05\cdot\left[M-n\right])$ on level completion, where $n$ is the number of steps taken and $M$ is the move limit.
    \item $+0.1$ for each power piece combination used.
\end{itemize}

\subsection{Prediction methods}
\label{sec:modelling methods}

We test out three different methods for this problem:
a neural network (\textbf{NN}) and a random forest (\textbf{RF}) approach which generally work well with dense data, and a factorisation machines (\textbf{FM}) method which extracts a latent representation of interactions that enables modelling sparse, high-dimensionality data such as user-item interactions.

\subsubsection{Random forest}

Random forest is an ensemble prediction method that utilises multiple decision trees that are created using a random subset of the features and data.
This enables non-linear modelling and predictions, and it has been used for the same task in previous work \cite{kristensen2022uist}.

RFs do generally not work very well with high-cardinality data such as unique user identifiers and levels since splits on individuals are not generalisable and can easily lead to overfitting.
A way to improve the model's ability to generalise to new data is by limiting the maximum depth and number of trees and using a smaller subset of features for building the trees.

\subsubsection{Artificial neural networks}

Artificial neural networks are another way to model nonlinear processes.
Although they are a more black-box approach than RF, their performance is usually comparable to or better than RF methods, and they can be trained incrementally while keeping the model size constant.
This is unlike RF methods which either need to be fully retrained or incrementally retrained by adding new individual trees to the ensemble.

Artificial neural Networks can be prone to overfitting due to the large number of parameters in the models. In order to improve their ability to generalise, using regularisation and/or adding dropout layers to the model are commonly employed strategies~\cite{srivastava2014dropout}.

\subsubsection{factorisation machines}

Factorisation machines are a general-purpose method that can be used for both classification and regression tasks and belong to a family of matrix factorisation methods in which entities are described by an embedded latent vector that enables modelling of a sparse interaction between entities~\cite{rendle_factorisation_2012}.
Specific to FMs, the features can consist of high-cardinality data, such as user and level identifiers, and dense data such as content tags or level descriptors.
The formula is given by

\begin{equation}
    \hat{y} = w_0 + \sum_{i=1}^n{w_i x_i} + \sum_{i=1}^n\sum_{j=i+1}^n{\langle \mathbf{v}_i, \mathbf{v}_j\rangle x_i x_j},
    \label{eq:fm equation}
\end{equation}

\noindent where $w_0$ is a global bias, $w_i$ is the bias of the $i$'th variable, $\mathbf{v}_i = [v_1, \dots, v_k]$ is the latent representation of the $i$'th variable using $k$ latent factors, and $\langle \mathbf{v}_i, \mathbf{v}_j\rangle$ is the dot product between the $i$'th and $j$'th that captures the second-order interactions of the variables. 

Matrix factorisation methods, such as FMs, are particularly prone to \textit{cold start} problems: to learn a latent representation of a given item, it needs to be present in the training data, but that is not possible with unreleased content (e.g. new levels in puzzle games).
One possible way to deal with this in FMs is by including additional data, such as tags or other item features, which is possible to learn latent representations from. 

\section{Data}
\label{sec:data}

\begin{table}[ht!]
\centering
\begin{tabular}{p{2.7cm}p{3cm}p{5.2cm}}
\textbf{Type of feature}                               & \textbf{Name} & \textbf{Description}                              \\ \hline
\multicolumn{1}{c|}{\multirow{2}{*}{Historic data}}  & Average attempts (level) & Average attempts on the level based on playthroughs by old players \\
\multicolumn{1}{c|}{}                                  & Average attempts (player) & Average attempts by the player on the first 100 levels \\[0.5mm] \hline
\multicolumn{1}{c|}{\multirow{4}{*}{Player features}}  & Moves used & Number of moves used relative to the move limit when completing the level \\
\multicolumn{1}{c|}{}                                  & In-game boosters & Boosters that can be used during the level\\
\multicolumn{1}{c|}{}                                  & Pre-game boosters & Boosters that can be used before starting the level \\[0.5mm] \hline
\multicolumn{1}{c|}{\multirow{6}{*}{Level attributes}} & Board size & Number of available board slots \\
\multicolumn{1}{c|}{}                                  & colour entropy & Entropy of colour spawning weights; $S=-\sum_{i}{p_i\log p_i}$ \\
\multicolumn{1}{c|}{}                                  & colours & Number of unique colours in the level \\
\multicolumn{1}{c|}{}                                  & Board pieces & Multiple features that each describe the number of board pieces on the initial board \\
\multicolumn{1}{c|}{}                                  & Collect goals & Multiple features that each describe the number of collectgoals \\[0.5mm] \hline
\multicolumn{1}{c|}{\multirow{7}{*}{Agent features}} & Training steps & Number of training steps until minimum average length was achieved \\
\multicolumn{1}{c|}{}                                  & Move limit & Move limit of the specific level \\
\multicolumn{1}{c|}{}                                  & Game length \textit{(min + std)} & Average number of moves used to complete level across 8 seeds \\
\multicolumn{1}{c|}{}                                  & Completion rate \textit{(min + std)} & Fraction of times the agent finished the level within the movelimit/100 moves across all 8 seeds \\
\multicolumn{1}{c|}{}                                  & Action entropy \textit{(min + std)} & RL agent action entropy \\
\multicolumn{1}{c|}{}                                  & Training losses \textit{(min + std)} & RL policy and value losses \\[0.5mm] \hline
\end{tabular}
\caption{Features investigated for the different models. The player features are aggregated means on the first 100 levels.
The values of the agent features are averaged from the last 100 episodes/playthroughs. For some of the agent features, we include both the value at the time step where the smallest average number of steps was achieved (denoted \textit{min}), and the standard deviation of the feature between time steps 1M to 9M (denoted \textit{std}).}
\label{tab: supervised features}
\end{table}

The exact type of data available will depend on the game. However, we can define four categories that the data can belong to, following the approach by Wheat et al.~\cite{wheat2016modeling}:

\begin{itemize}
    \item \emph{Historic data} that consider the average number of attempts by previous players, per level or per player basis.
    \item \emph{Player data} that capture aspects related to player skill and purchase tendency.
    \item \emph{Level data} that capture specific types of game mechanics and descriptors of the level.
    \item \emph{Agent data} that captures more dynamic information about the level and is extracted using a playtesting method.
\end{itemize}

Each of these categories will have an impact on the performance of the prediction model, but they may not be fully informative on their own.
For example, player data is necessary to include for personalised predictions, but on its own, it does not say anything about the level itself or how easily the player can handle specific gameplay elements.
Similarly, agent data may reveal some intrinsic difficulty/ground truth pass rate, but without level data, it is hard to account for biases in the agent algorithm (i.e. certain mechanics may be easier to learn for an agent compared to a human~\cite{gudmundsson2018human}), and personalised predictions are impossible without player data.
Lastly, historic data may be informative, but they are not available on new content or for new players.
We, therefore, experiment with combining the data in a number of ways to test the importance of each data category.
We consider the following combinations:

\begin{itemize}
    \item \textbf{Historic:} Only player-level-attempt information. For FMs, this is implicitly learned in the variable biases, while for the RF and NN approaches, it needs to be calculated explicitly. Only available for predictions on existing content.
    \item \textbf{Player+level:} Both player and level data available for personalised predictions.
    \item \textbf{Agent:} Only agent data available used for non-personalised predictions.
    \item \textbf{All:} Player, level, and agent data combined, possible to use for non-personalised or personalised predictions. It does not include features that are explicitly derived from historic observations.
\end{itemize}

\subsection{Lily's Garden case study data}

The data used in this study consists of the number of attempts each player has spent on a given level and is recorded between 2021-06-01 and 2021-12-01.
Only players who have started playing the game in this interval and played at least 200 levels are included in the data set.
We use data that come from the four different categories of data mentioned previously. 
An overview of these features is shown in Table \ref{tab: supervised features}.

The player data consists of aggregate player data over the first 100 levels.
We consider the average number of moves they have used to complete the levels, as well as booster usage on the first 100 levels.
These features capture the general competence of the players and their willingness to pay to progress, which are both strong predictors for how many attempts they will use on future levels~\cite{kristensen2022uist}.

The level attributes describe the size of the board and the goals and board pieces present in the level.
Additionally, the number of colours and the entropy of the colour distribution are also included, which can affect how easy it is to make stronger combinations and thus the difficulty of the level.
Unlike previous work where some of the level features consisted of historic data such as the average number of attempts on the level, for this study, we only include features that are possible to determine before any players played the given level.
This is necessary since in the experiments we consider the scenario with new levels where no historic data is available.

Lastly, for the agent data, we include information about how well the agent learned to play the level and how easily it learned to do so.
We do this by recording some features at the point during training at which the average number of moves taken by the agent is the lowest, as well as the standard deviation of the features between the time steps 1M to 9M.

\section{Experiment A: personalised predictions}
\label{section:experiment 1 personalised predictions}

\begin{figure}[t]
    \centering
    \includegraphics[width=0.95\columnwidth]{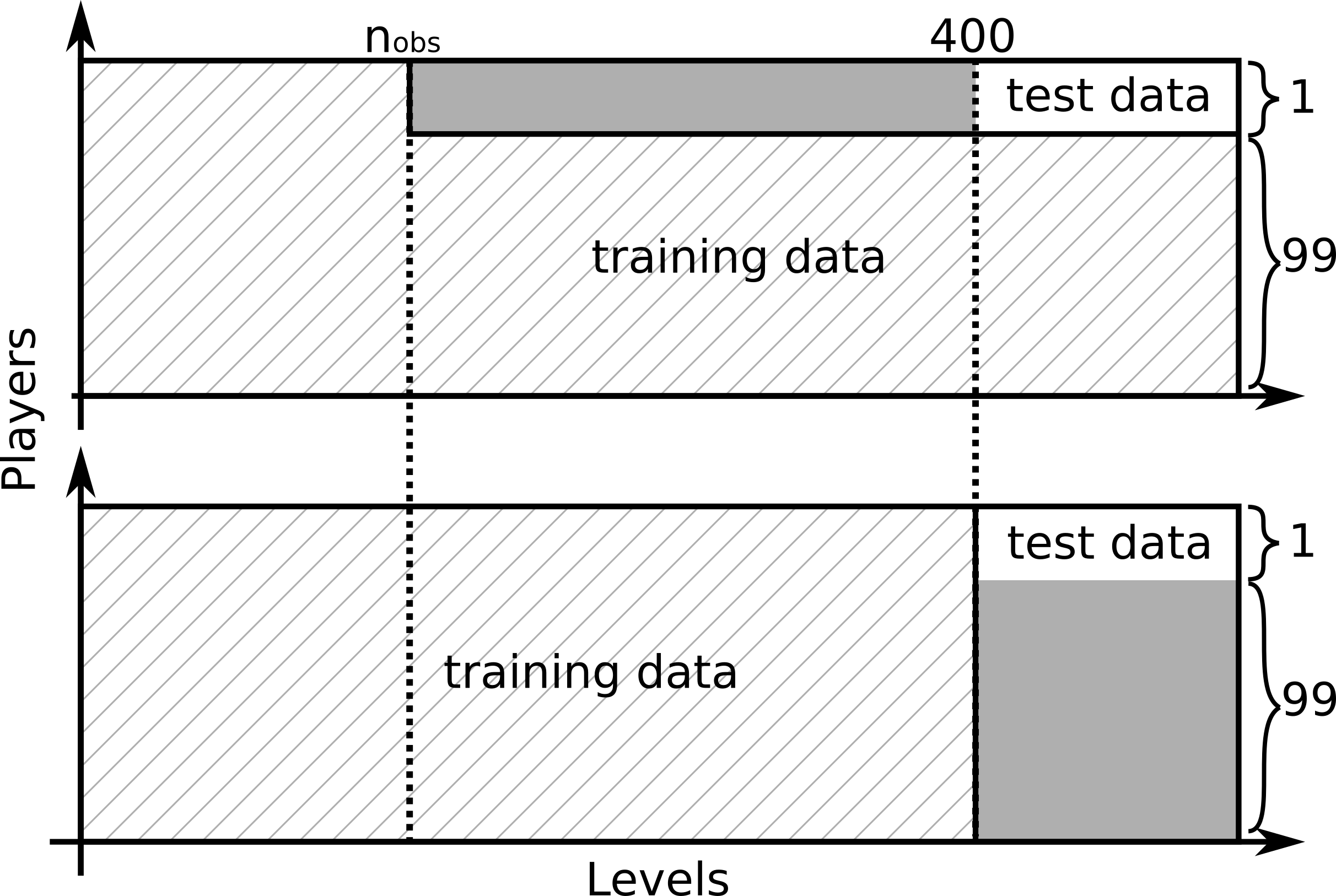}
    \caption{Train/test split of the data for personalised predictions on levels with historic data and on levels in the cold start scenario.
    In this study we use $n_\textup{obs}=100$. The greyed out area is ignored observations for ensuring the experiments are evaluated on the same content. Adapted from \cite{kristensen2022uist}.}
    \label{fig:training split}
\end{figure}

The first research question (RQ1) is aimed at investigating the modelling of the perceived difficulty of a puzzle level for a specific player.
We follow a similar approach as Kristensen et al.~\cite{kristensen2022uist} where we use data from the commercial puzzle game, Lily's Garden, which is described in Section \ref{sec:case study}.
However, we extend the study by also considering a cold start scenario where there are no observations available for a given level.
We denote the two cases as \textit{playthrough/historic data} (PD) and \textit{cold start} (CS).

To create data sets for training and testing, we split the data in two different ways, which are shown in Fig. \ref{fig:training split}.
In the first case, we split the players into train/test sets with a 99-1 split but include observations of the players from the test set in the training set-up until the first $n_\textup{obs}$ levels.
In the second case, we split by level number, where the first 400 levels are used for training and the last 99 levels are used for testing.

The nature of the prediction task, in which very difficult levels can have a large variance in attempts, as shown previously in Fig. \ref{fig:attempt distribution example}, means that the best-performing models in terms of root mean squared error (RMSE) and mean absolute error (MAE) may be different.
Since the models are optimised using RMSE and we want the method to be more robust towards more difficult levels, we choose to use RMSE as the reported metric.
We calculate it from the same group of players above level 400 in all cases.

We test out the three methods described in Section \ref{sec:modelling methods}, and we train them using the four different combinations of data outlined in Section \ref{sec:data}.
In the first case, we include a naive baseline prediction for each level which is calculated as the average number of attempts on the given level by the players in the training set.
In the second case, we use a constant prediction of $y_\textup{baseline} = 3.23$ that is calculated as the average number of attempts on levels 100-400.

\begin{figure}[t]
    \centering
    \includegraphics[width=0.95\columnwidth]{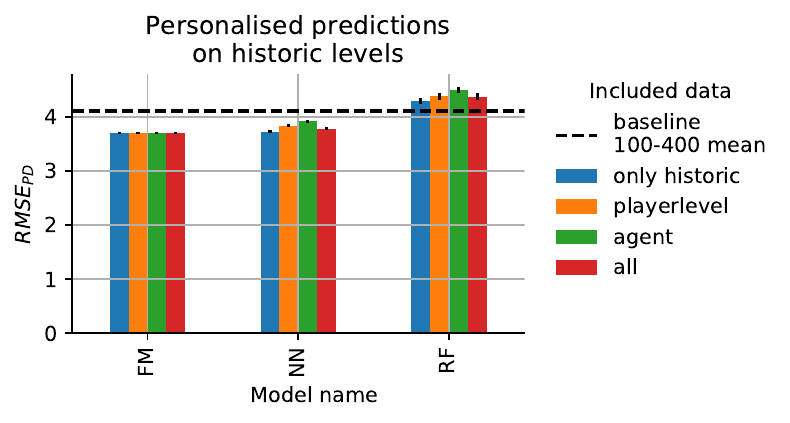}
    \caption{RMSE of personalised predictions on content that has been played by other players.}
    \label{fig:personalised existing content}
\end{figure}

\subsection{Personalised predictions with historic data available}
The performance of the methods in terms of RMSE in the first case is shown in Fig \ref{fig:personalised existing content}. 
The best performing model is the FM model which only includes the player and level data. 
FMs appear to be more robust in identifying possible bottlenecks for an individual player, where they possibly require a large number of attempts and may feel stuck.
This can be used to the game designers' advantage by preemptively assisting the players if, for example, the estimated perceived difficulty exceeds a certain threshold.

Another observation to note is that, while the NN and FM approaches perform better than the baseline, the RF model appears to perform worse than all other methods. 
A crucial difference between this study and the study by Kristensen et al. (2022) \cite{kristensen2022uist} is that the level features do not include the historic average attempts because this information would not be available when estimating the difficulty of unreleased levels.
Without such a strong predictor, the RF model does not appear to be able to infer a meaningful average prediction.
Furthermore, both the FM and NN approaches do not employ average attempts as input, but they may be able to infer the average attempts on each level through either the complexity of the network or the variable biases, $w_i$, of the FM model.

\begin{figure}[t]
    \centering
    \includegraphics[width=0.95\columnwidth]{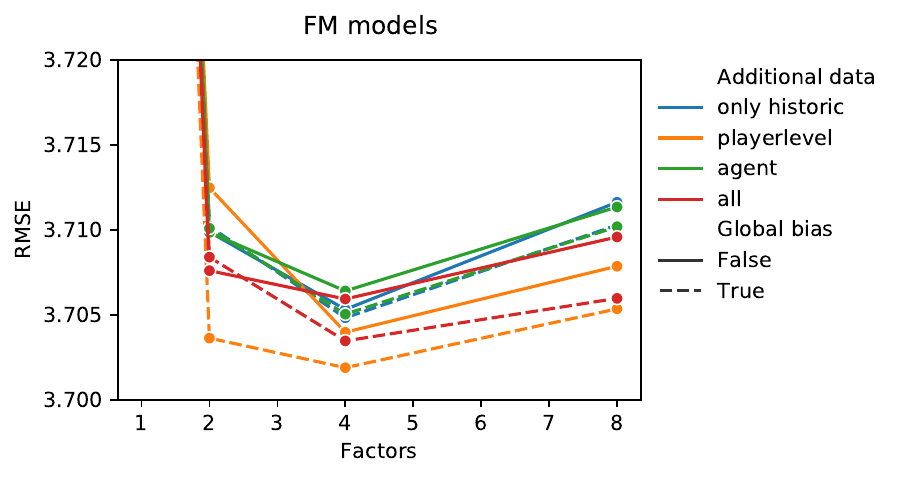}
    \caption{RMSE of the tested FM models for personalised predictions on current levels as a function of number of factors. The colours indicates what additional datba that are included in the data set. The dashed line indicates whether $w_0$ is included in the model or not (see Eq. \ref{eq:fm equation}). }
    \label{fig:personalised modelling errors}
\end{figure}

The impact of using different types of data as input to the FM method is shown in Fig. \ref{fig:personalised modelling errors}. 
The results are in line with previous studies~\cite{kristensen2022uist} in which the RMSE and MAE are around $3.8$ and $2.3$, respectively, and augmenting the data with player and level data leads to a minor improvement.
An extension to the previous study also includes simulated agent data; this data does not significantly improve the predictions for the FM method, and the addition leads to worse performance in combination with the player and level data categories. 
A closer inspection of the training process shows that the performance on the test set worsens after just 50 iterations when including all data categories, while in the other cases, a performance drop happens after around 950 iterations.
This suggests that a likely cause for the poorer performance is the tendency of the model to overfit.
Furthermore, in a high-dimensional space with strongly correlated parameters, the Gibbs sampler used in the MCMC optimiser for the FM method may not converge~\cite{justel1996gibbsmasking}, which is the case with the colour entropy and colours of the level attributes, and the minimum and standard deviation of the games duration.

In terms of other model parameters, four factors seem to work best for optimising RMSE but at the cost of increasing the MAE.
However, the difference in performance when using between two and four factors is not significant and does not contradict the results of previous works in which two factors were concluded to be sufficient.
Using additional data does not warrant the use of more than four factors and can, in fact, worsen the performance.
There is no clear pattern of whether using a global bias improves the predictions, although the best-performing model does include a global bias.

\subsection{Predictions in cold start scenarios}
\label{subsec:experiment 1 cold start}

In the second case, the performance of the models is very similar as seen in Fig. \ref{fig:personalised existing content}.
Excluding the models with only historic data, since this is not available in the cold start scenario, all three FM models perform better than the comparable NN and RF models.
Similarly, only the RF models perform worse than the constant baseline.
The same parameters for the FM methods (4 factors, both global and variable bias included) lead to the best performance of the tested FM configurations.

\begin{figure}[t]
    \centering
    \includegraphics[width=0.95\columnwidth]{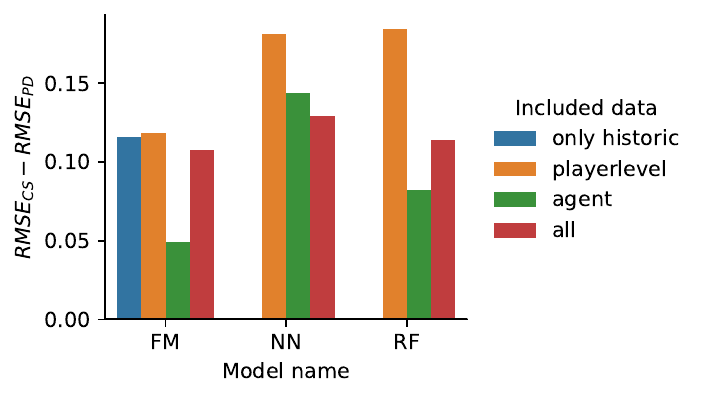}
    \caption{Difference in RMSE for personalised predictions between the scenario where data about player playthroughs (\textbf{PD}) is available and the cold start (\textbf{CS}) scenario. A positive difference indicates that the error in the cold start scenario is higher.}
    \label{fig:personlised rmse difference}
\end{figure}

Since the personalised predictions in both cases have very similar RMSE, we look at the difference between the errors in the two cases in Fig. \ref{fig:personlised rmse difference} to better compare the performances.
The results show that for all the methods, the error in the cold start scenario is higher regardless of data type.
This is not surprising since the information from other players' performance on a level should be a strong predictor of the perceived difficulty of other players.
In the FM approach, including agent data does lead to a smaller gap in error in the two scenarios, which suggests that agent data does contain important predictors for predicting difficulty that the FM model learns to utilise.
This is also true for the other methods, but the difference between the two scenarios for the NN and RF model is still larger than in the FM case.

It is notable that the performance of the models is similar and that the FM method with no included data performs better than a constant baseline.
Since the FMs reduce to a simple linear function that consists of a global bias and bias term for the players according to Eq. \ref{eq:fm equation} in cold start scenarios, it suggests that the learnt biases for individual players are sufficient to capture the general performance of the players on both existing and new content.
This is valuable information for game designers, as it may be possible for them to use these learnt biases for more individualised content.

\section{Experiment B: Cohort predictions}

The second research question is regarding the possibility of making cohort level predictions.
Level designers are typically interested in knowing the average number of attempts on the levels, and since different cohorts reach the levels at different times, the measured average difficulty may change over time, depending on the cohort.
This experiment simulates this by comparing the aggregated average predictions per level with the average attempts on each level for the cohort in the test set.
Similar to the previous experiment, we consider both the scenario where there is historic data from previous players' playthroughs on a given level and a cold start scenario.
We use the same baselines as in the previous experiment.

\begin{figure*}[t]
    \centering
    \includegraphics[width=\columnwidth]{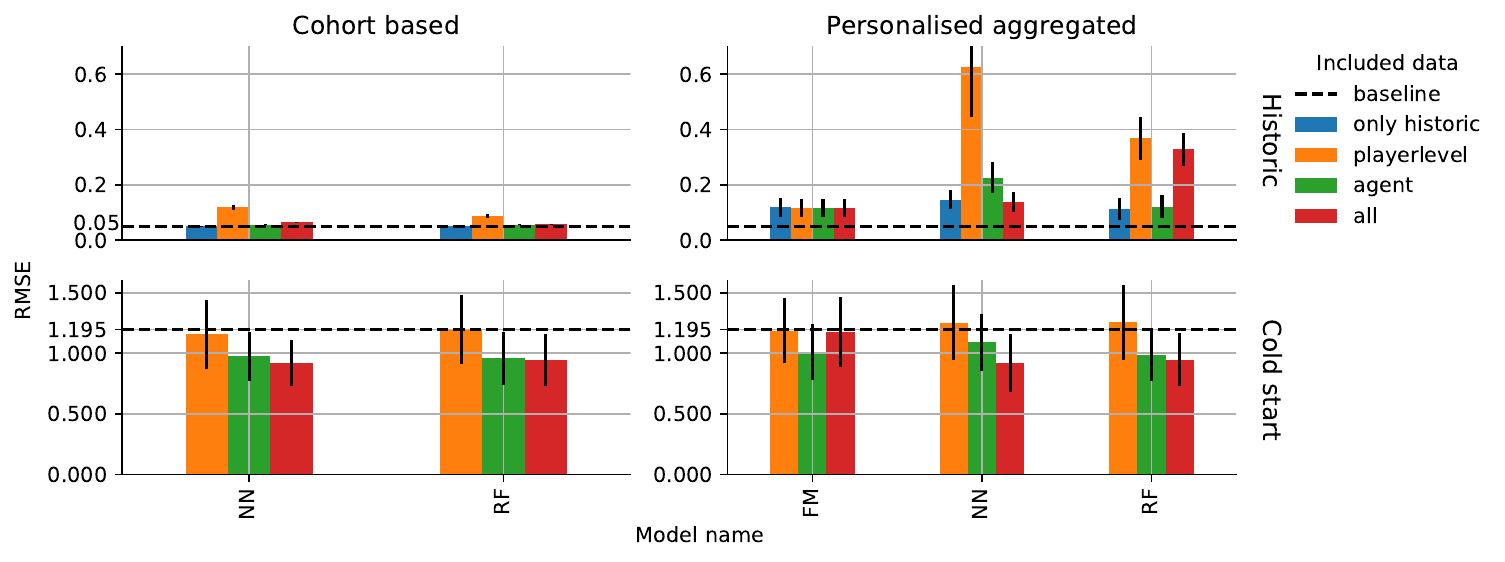}
    \caption{RMSE of the different approaches for predicting the average difficulty of a level for a cohort. The left plots show the cohort-based per-level predictions, and the right plots show the aggregated average of the personalised predictions for the full data set. The top row is in the case where observations on the examined levels exist, while the bottom row is the cold-start scenario. Note that historic data is not available in the cold start scenario.}
    \label{fig:prediction errors}
\end{figure*}

We consider two ways of predicting the average number of attempts on a cohort level.
The first approach is using the personalised predictions from the first experiment and computing the aggregated average on each level.
For the second approach, since individual estimates may be noisy or inaccurate, we group the players into 10 random cohorts and use the aggregated average values of the player features to describe these cohorts.
The regression target for each cohort is then their average number of attempts on the given level.
This allows us to directly use the tested methods to predict the average number of attempts on a per-level basis of the cohort of  players in the test set.
These two approaches will be denoted by how the initial predictions are grouped (i.e. \textbf{personalised} or \textbf{cohort}) but we note that in both cases, the final estimates are on a cohort level.

\begin{figure}[b!]
    \centering
    \includegraphics[width=0.95\columnwidth]{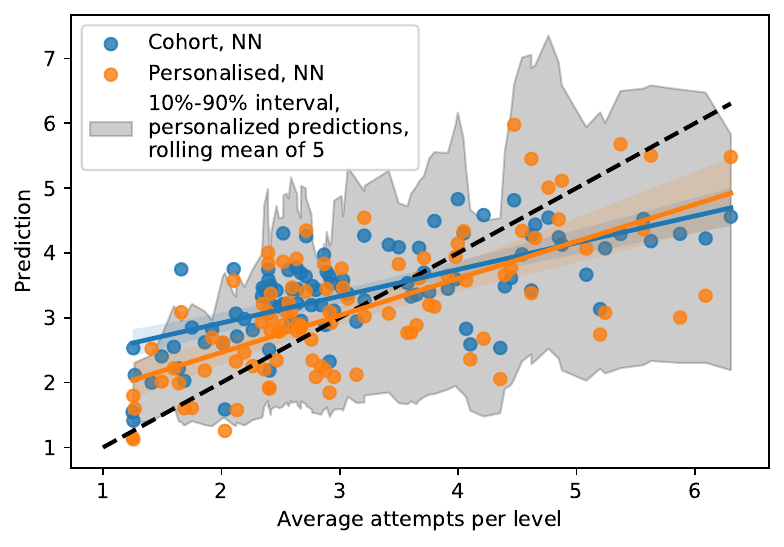}
    \caption{Example of predictions of the NN method with cohort and personalised predictions using all categories of data in the cold start scenario.}
    \label{fig:prediction scatterplot}
\end{figure}


A summary of the results achieved by the investigated methods is shown in Fig. \ref{fig:prediction errors}.
The errors for all models when predicting the difficulty of existing content is very low, which is not surprising since the cohorts are generally not very different as demonstrated by the very small RMSE of the baseline.
The personalised errors are generally larger than the cohort-based predictions, which is most likely due to the fact that these models are optimised to minimise the RMSE for each player.
The optimal model, in this case, is not necessarily one that captures the average number of attempts by the whole cohort, but rather one that can account for the high within-level variance.
This is clear for the FM models, in which using only one factor leads to an error similar to the baseline, but more factors lead to higher RMSE in this experiment where we show the results from the 4-factor models.

In the cold start scenario, both the NN and RF models are the best performing models and appear to have similar performance for both personalised and cohort predictions.
Additionally, including all possible categories achieves the best results for the NN and RF methods, similar to what was found in the first experiment.
This is not the case for FM with all data included, though, which seems to suffer from the same type of overfitting seen previously.
The model parameters for the best performing FM models were similar as well, where including both a global and variable bias as well as using 2-4 latent factors lead to the best performance.

Generally, including only player and level data is not sufficient for the model to perform better than the baseline. Only with the inclusion of agent data, it is possible to have more accurate predictions than a simple constant average.
By looking at the feature importances for the RF models, we also find that the agent features are on average 5-6 times more impactful than the level features, with the minimum game length and completion rate being among the most important.
These results highlight the importance of having a playtest agent that can capture dynamic aspects of a level to allow more accurate predictions in cold-start scenarios.
It also supports the conclusion by Gudmundsson et al.~\cite{gudmundsson2018human} that combining agent performance with level features is useful in order to be able to account for differences in how players and AI agents learn and behave.

Another notable result is that cohort predictions are on-par with personalised predictions for estimating the average level's difficulty in the cold-start scenario.
To visualise the differences between the two approaches, Fig. \ref{fig:prediction scatterplot} shows the predictions from each method compared to the actual average number of attempts.
Both methods tend to overestimate the difficulty of easy levels and underestimate the difficulty of the hard levels and tend to predict the mean number of attempts.
This suggests that they are both underfitted or that the features are not informative enough to generalise to new levels.
The personalised predictions tend to have a larger variance, especially at higher difficulties, which can be seen from the 80\% prediction interval.
This is not completely unexpected due to the high variance in attempts between players on harder levels, as shown in Fig. \ref{fig:attempt distribution example}.

\section{Discussion}

From the results of the experiments, it is clear that the tasks of creating personalised difficulty predictions for players on content with and without available player playthrough data require different approaches.
While a part of this is related to generalisability and overfitting, a deeper discussion about the data and models can help understand in which circumstances one approach works better than another.
In this section, we therefore focus on three topics: the difference between models for old and new content, the feasibility of using any of these models in practice and playtest agent data. 


\subsection{Choice of model}

One complication with a combined difficulty framework that can work on both old and new content is that the available data for the two scenarios are inherently different.
For new content, there are no direct observations of how many attempts actual players spend on a level.
This \textit{cold-start problem} can especially be an issue for factorisation-based approaches, such as FMs, since the bias and latent factors associated with new content can not be properly inferred with no observations.
While the inclusion of additional data in FM methods, such as tags~\cite{juan2016field} or player/level/agent data, can alleviate this problem, we still observe that the predictions tend to stay close to the global average and not capture the full range of behaviours of players on new levels.

For the RF and NN approaches, the cold-start problem does not exist to the same extent. FMs learn latent descriptions for each item, but the RF and NN methods instead only use the included data features.
This makes these latter models not rely on modelling specific items, at the cost of not having as accurate personalised predictions, as seen in the first experiment
It is therefore also not surprising that the difference between personalised and cohort predictions on new content is insignificant with those two methods: the features with high predictive power (mainly agent data) are the same, so the information that is possible to extract is the same.
Data for the personalised model training are essentially an oversampled dataset of the full dataset.

It is worth noting that, for the RF method, while the results of the second experiment for both personalised and cohort predictions were promising, the results from the first experiment showed that the RF method performs worse than the baseline to predict individual perceived difficulty.
The inability of the RF method to capture individual differences may be due to the greedy nature of the algorithm and the fact that the given player's features were too uninformative compared to both agent and level data.

Ultimately, both regarding training and the ability to capture complex behaviours, a neural network based approach is more feasible for personalised predictions on both old and new content, though at the cost of interpretability. For accurate and interpretable methods on existing content, the FM based approached appears more suitable.


\subsection{Difficulty modelling in practice}

An important question about this approach is whether these results are reflective of how these methods would work in practice. To answer that, we consider some aspects regarding how we collect and split the data.

The data used in this study was collected over a 6 month period, but there is no information about when the players started or stopped playing or which cohort they might belong to.
In order to enable the level designers of a live game to be proactive, a more realistic approach would be to consider whether the player has been active recently or not.
While the approach used in this study made it possible to compare it with previous work (Kristensen et al. (2022) \cite{kristensen2022uist}), an idea for a future study is to use a rolling window where the training data only include data before a given date, and the test data only includes after this date up to the prediction date where the level designers would use the results.

Another consideration regarding the way the data is split is whether the scenario of having historic data or the cold start scenario is necessary.
While the cold start scenario is representative of completely new content, there are often other concerns when creating new levels, such as whether the puzzle \textit{feels} fun or if the solution(s) are clear.
This is a much harder problem to solve and requires the level designers to play through the levels regardless.
This also means that, in practice, there is a smooth transition from a complete cold start scenario to a scenario with \textit{some} historic data available -- and not just by the level designers but also by other players due to pre-releases to certain regions or A/B tests.
Since FMs should work well with such sparse data, an interesting line of research would therefore be to investigate how many observations of attempts on a level are necessary to beat the baseline.
This question of how many observations are necessary has been explored by Kristensen et al.~\cite{kristensen2022uist} for players, but it would be valuable to learn whether the results are true for levels too.

We also note that, to address this cold start scenario, we experimented with treating the agent as a player.
It is a way that could help alleviate the cold start problem in FMs where enriching item data by adding new ratings (e.g. \cite{tahmasebi2021hybrid}) can augment very sparse observations.
However, initial experiments did not show promising results, and an even stronger tendency to predict the global average was observed.

With the ultimate goal of being able to employ this difficulty framework in a live game, the main concern is whether the accuracy on especially new content is accurate enough for the level designers to rely on.
Gudmundsson et al.~\cite{gudmundsson2018human} consider the same business problem and present a method based on a similar two-step approach that employs a playtest agent to extract an agent pass rate and subsequently use in a prediction model.
They mentioned that King has used it for two of their games and reported the best model has a 4.0\% to 6.6\% MAE on the pass rate.
Translating our results from attempts to pass rate, our method has a $8.1\%$ MAE.
However, it is not possible to translate the result directly since they do not report the absolute magnitude of the win rates, which means if they have more hard levels with low, but similar, pass rates, the reported MAE is also expected to be lower.
Additionally, they considered two different games which share many similar mechanics but still observed a difference in the predicted performance.
While it is therefore hard to make a direct comparison, the results in this study can still inform the level designers about the scale of difficulty, and further ideas could involve classifying the levels into easy, medium, and hard difficulties rather than directly predicting the number of attempts. 
Furthermore, the variation of the results within the experiment by Gudmundsson et al.~\cite{gudmundsson2018human} and between that and this experiment, albeit minor, can be further studied to investigate whether there are specific differences in the player base or the characteristics of the games or the methods employed that lead to the observed results.






\section{Conclusion}

In this study we considered two questions:
is it possible to estimate the perceived difficulty of a puzzle level for a given player on both existing newly generated (human or otherwise) levels, and is it possible to do this on a cohort level?
To answer these questions, we tested three different prediction methods to estimate the number of attempts each player is expected to use in order to complete a given level using data from a live commercial puzzle game.
Additionally, we experimented with including and combining data belonging to four categories: historic data, player and level data, as well as agent data collected by a reinforcement learning playtesting agent.

For personalised predictions on existing content, with previous players' data, the factorisation machines approach performed significantly better than the random forest and neural network methods. 
However, including all three types of data for the factorisation machine led to worse performance most likely due to overfitting and bad optimisation conditions.
The neural network approach worked better than the baseline, while the random forest performed worse.
This is explained by the fact that the NN method is able to capture the historic average number of attempts through the complexity of its network and sequential training that enables better handling of large data sets.

To estimate the average difficulty of new content, in a cold start scenario, the neural network approach worked the best for both personalised and cohort-based predictions, while the factorisation machines method did not work very well due to the cold start problem common in factorisation methods.

It was also found that including the data collected by a playtest agent is necessary for the methods to perform better than the global average.
However, since the agent consistently performed better than most players, it did not differentiate easily between the easy to medium difficulty levels.

\bibliographystyle{IEEEtran}
\bibliography{IEEEabrv,bibliography}

\end{document}